\documentclass{article}

\usepackage{microtype}
\usepackage{graphicx}
\usepackage{subcaption}
\usepackage{booktabs} 

\usepackage{hyperref}

\usepackage[preprint]{icml2026}

\usepackage{amsmath}
\usepackage{amssymb}
\usepackage{mathtools}
\usepackage{amsthm}
\usepackage{multicol}
\usepackage{multirow}
\usepackage[capitalize,noabbrev]{cleveref}

\theoremstyle{plain}

\theoremstyle{definition}

\theoremstyle{remark}

\usepackage[textsize=tiny]{todonotes}

\icmltitlerunning{Unleashing MLLMs on the Edge: A Unified Framework for Cross-Modal ReID via Adaptive SVD Distillation}

\begin{document}

\twocolumn[
  \icmltitle{Unleashing MLLMs on the Edge: A Unified Framework for Cross-Modal ReID via Adaptive SVD Distillation}



  \icmlsetsymbol{equal}{*}

  \begin{icmlauthorlist}
    \icmlauthor{Hongbo Jiang}{equal,sch}
    \icmlauthor{Jie Li}{equal,sch}
    \icmlauthor{Xinqi Cai}{sch}
    \icmlauthor{Tianyu Xie}{sch}
    \icmlauthor{Yunhang Shen}{comp}
    \icmlauthor{Pingyang Dai}{sch}
    \icmlauthor{Liujuan Cao}{sch}
  \end{icmlauthorlist}

  \icmlaffiliation{comp}{Tencent Youtu Lab, Shanghai, China}
  \icmlaffiliation{sch}{Xiamen University, Media Analytics and Computing Lab, Department of Artificial Intelligence, School of Informatics, Xiamen, China}

  \icmlcorrespondingauthor{Liujuan Cao}{caoliujuan@xmu.edu.cn}
  \icmlkeywords{Person Re-identification, Multimodal Large Language Models, Knowledge Distillation, ICML}

  \vskip 0.3in
]



\printAffiliationsAndNotice{}  

\begin{abstract}
Practical cloud-edge deployment of Cross-Modal Re-identification (CM-ReID) faces challenges due to maintaining a fragmented ecosystem of specialized cloud models for diverse modalities. 
While Multi-Modal Large Language Models (MLLMs) offer strong unification potential, existing approaches fail to adapt them into a single end-to-end backbone and lack effective knowledge distillation strategies for edge deployment.
To address these limitations, we propose MLLMEmbed-ReID, a unified framework based on a powerful cloud-edge architecture. 
First, we adapt a foundational MLLM into a state-of-the-art cloud model.
We leverage instruction-based prompting to guide the MLLM in generating a unified embedding space across RGB, infrared, sketch, and text modalities. This model is then trained efficiently with a hierarchical Low-Rank Adaptation finetuning (LoRA-SFT) strategy, optimized under a holistic cross-modal alignment objective.
Second, to deploy its knowledge onto an edge-native student, we introduce a novel distillation strategy motivated by the low-rank property in the teacher's feature space. To prioritize essential information, this method employs a Principal Component Mapping loss, while relational structures are preserved via a Feature Relation loss.
Our lightweight edge-based model achieves state-of-the-art performance on multiple visual CM-ReID benchmarks, while its cloud-based counterpart excels across all CM-ReID benchmarks. The MLLMEmbed-ReID framework thus presents a complete and effective solution for deploying unified MLLM-level intelligence on resource-constrained devices.
The code and models will be open-sourced soon.
\vspace{-2mm}
\end{abstract}


\section{Introduction}
\label{sec:intro}
\begin{figure}
  \centering
   \includegraphics[width=\linewidth]{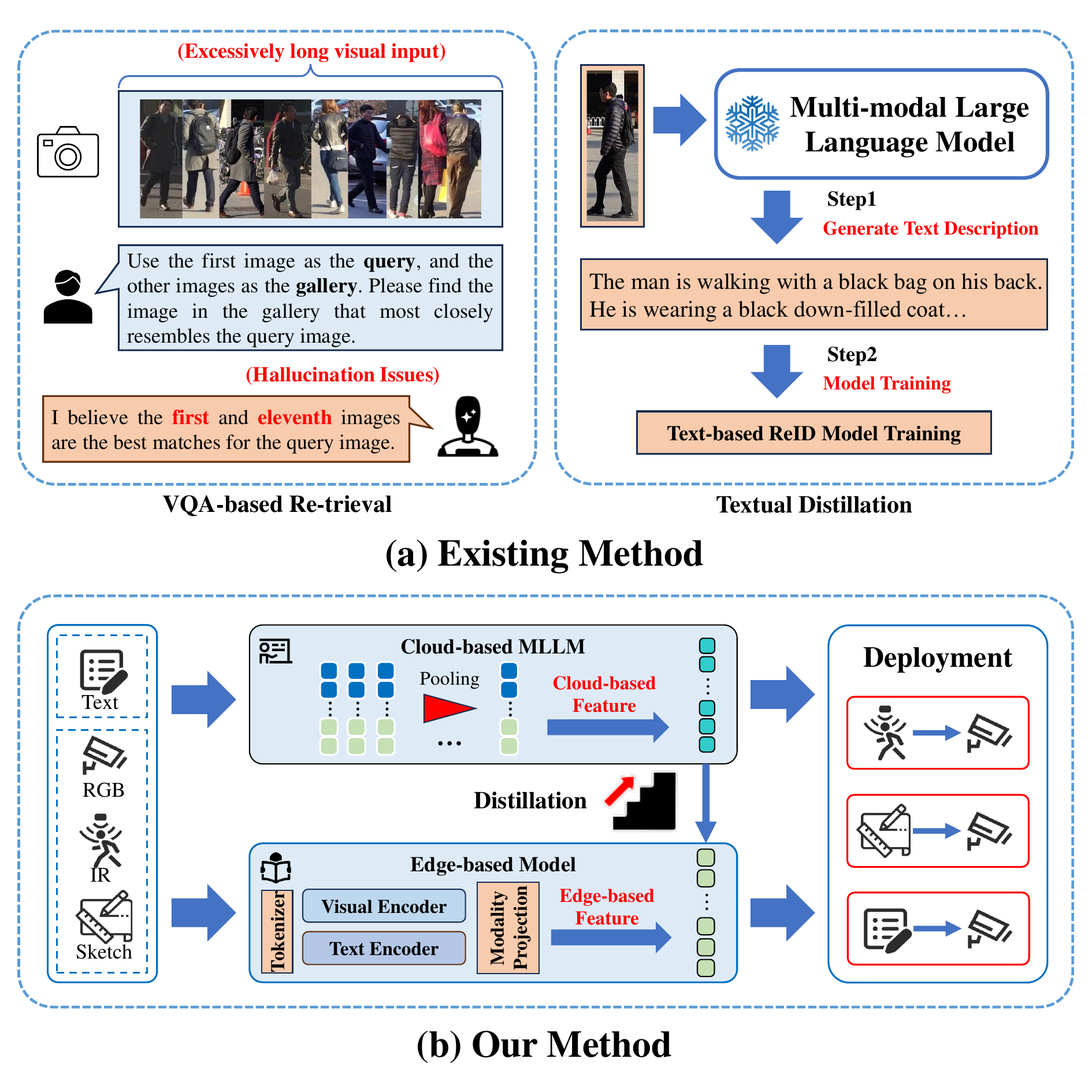}
   \caption{A Comparison of MLLM-based ReID paradigms. Images are from the QrCM-ReID dataset. (a) Existing methods indirectly apply MLLMs for VQA-based retrieval (limited by gallery size, prone to hallucination with long visual contexts) or textual distillation (restricting to text-only ReID). (b) Our MLLMEmbed-ReID directly uses a cloud-based MLLM as a unified teacher for diverse modalities. Its unified knowledge is distilled to a lightweight edge student for practical deployment.}
   \label{fig:motivation}
   \vspace{-2mm}
\end{figure}

Person re-identification (ReID)~\cite{sun2024comprehensive,wu2022overview} is a fundamental computer vision task for applications ranging from intelligent surveillance to public safety. While single-modal ReID (SM-ReID)~\cite{he2021transreid} has matured, real-world systems increasingly require Cross-Modal re-identification (CM-ReID)~\cite{jiang2023cross,chen2022sketch,zhang2023diverse} capabilities, such as matching infrared to RGB images. This necessitates learning the feature that is both discriminative and modality-invariant. Concurrently, the practical deployment of ReID has converged on a cloud-edge collaborative architecture~\cite{gu2023ai,wang2024end}. In this paradigm, high-capability models on cloud servers perform large-scale retrieval, while the edge models provide low-latency inference. A critical challenge in this setting is the scalability of the cloud component, as maintaining a fragmented ecosystem of specialized, computationally-intensive models for each CM-ReID pairing is inefficient and unsustainable.


Multimodal Large Language Models (MLLMs)~\cite{wang2025qwen2vl,yang2024mllmreid,lu2025llava,niu2025chatreid,bai2025qwen2_5}, are strong candidates to resolve this by serving as a singular, unified cloud model. Prior research has explored integrating MLLMs, such as generating textual descriptions for text-based ReID or reformulating the task into a Visual Question Answering (VQA) problem.
As depicted in \Cref{fig:motivation} (a), these existing paradigms are often indirect and suffer from practical limitations such as context length constraints and factual hallucination, or are confined to single text-based ReID tasks through textual distillation.
While innovative, these approaches do not employ the MLLM as a singular, end-to-end feature extractor for diverse CM-ReID tasks. The full potential of MLLMs as a truly unified backbone thus remains underexplored.
Furthermore, MLLMs' immense size creates deployment barriers for edge devices, and conventional knowledge distillation methods are ill-suited.
Experiments show conventional distillation approaches fail in the text-based ReID task.
At the same time, we found that traditional distillation methods treat MLLM's structured knowledge as a holistic black box, neglecting its internal feature attributes.


We introduce MLLMEmbed-ReID, a novel cloud-edge collaborative framework.
We first propose a novel and powerful cloud-based teacher model. By adapting a foundational MLLM using a hierarchical LoRA-SFT and a composite ReID training objective, we learn a single, shared embedding space for RGB, infrared, sketch, and text modalities without any task-specific modules. During inference, this teacher model processes multimodal data via instructional prompts and extracts a globally-aware ReID feature using a pooling operation on its final hidden states. This approach establishes a new state-of-the-art for unified CM-ReID at the cloud level, providing a robust foundation of knowledge for edge deployment.


Building upon this powerful teacher, we then tackle the critical challenge of edge deployment. We introduce a novel knowledge distillation strategy based on a key empirical observation: the teacher's ReID feature space exhibits a distinct low-rank phenomenon, with discriminative information concentrated in a small subset of principal dimensions. Leveraging Singular Value Decomposition (SVD), our method designs a structured learning curriculum that explicitly guides the lightweight student to master principal components first, while aligning feature correlations to indirectly learn vital minor dimensions. Extensive experiments demonstrate that this structured approach enables our lightweight edge model to achieve state-of-the-art performance on visual CM-ReID tasks.


Our main contributions are threefold:
\begin{itemize}
\item We pioneer the end-to-end adaptation of an MLLM into a singular, unified backbone for diverse CM-ReID tasks. Our method creates a new state-of-the-art cloud model by leveraging instruction-based prompting to generate a unified embedding space across four modalities, and coupling this with a hierarchical LoRA-SFT strategy under a holistic alignment objective for efficient and effective fine-tuning.
\item We discover the low-rank property in MLLM's ReID feature space and propose a novel distillation strategy that employs Principal Component Mapping and Feature Relation losses to structure knowledge transfer for efficient edge deployment.
\item We present a complete cloud-edge collaborative solution validated by comprehensive experiments, demonstrating that our lightweight edge model achieves state-of-the-art performance while maintaining the unified intelligence of MLLMs on resource-constrained devices.
\end{itemize}


\section{Related Work}
\label{sec:relatedwork}

\begin{figure*}
    \centering
    \includegraphics[width=\textwidth]{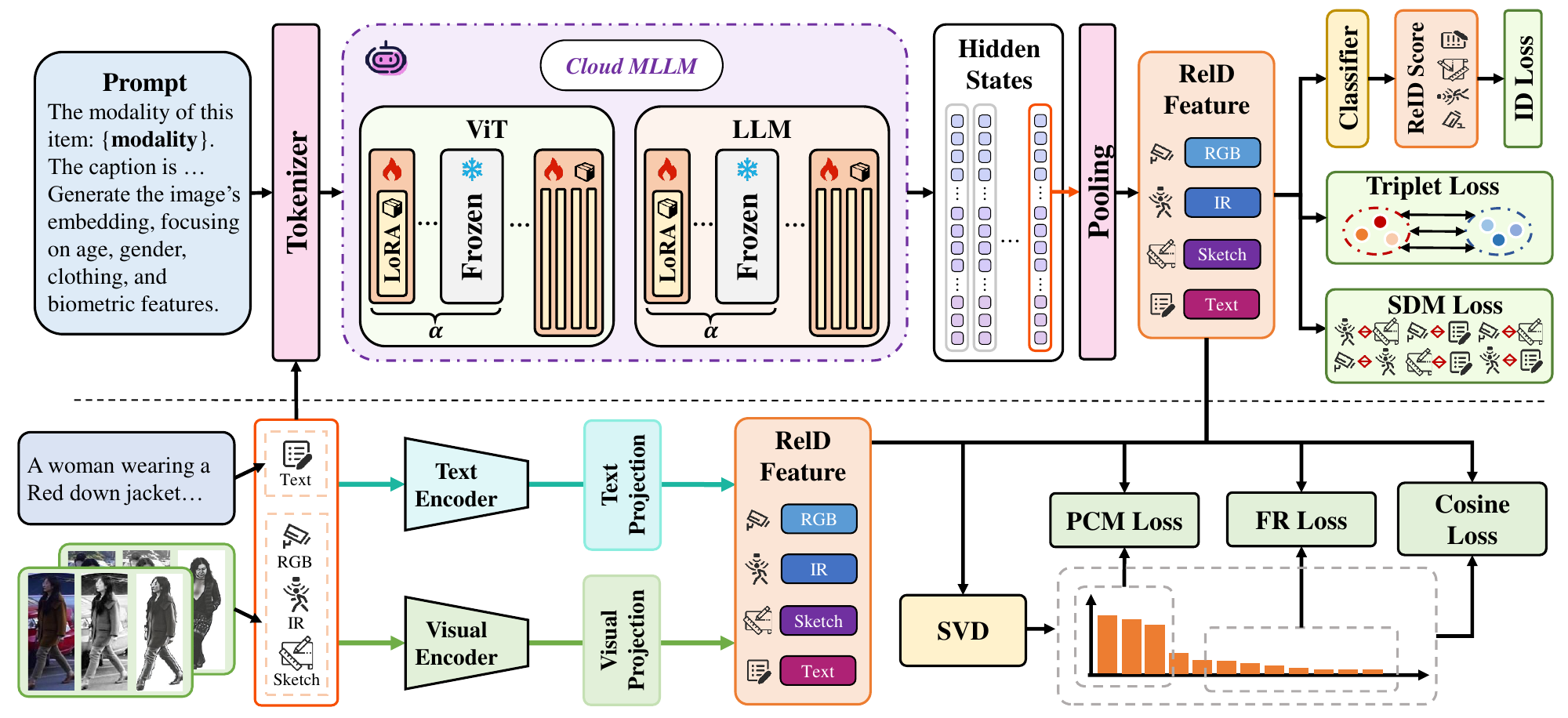}
    \caption{\textbf{Overview of the proposed MLLMEmbed-ReID framework}. Images are from the QrCM-ReID dataset. It primarily consists of two components: cloud model fine-tuning and edge model distillation. The cloud model includes task instructions and modality prompts, an MLLM backbone (Qwen2-VL), pooling operations, Identity Identification (ID loss), triplet learning (Triplet loss), and Similarity Distribution Matching (SDM). The edge model primarily includes Vision Language Model (VLM) backbone (CLIP (ViT-L/14)), modality projection, distillation matching loss (e.g., cosine loss), Principal Component Mapping Loss (PCM loss), and Feature Relation Loss (FR loss). Within the MLLMEmbed-ReID framework, both cloud and edge models can end-to-end unify the completion of CM-ReID tasks.}
    \label{fig:framework}
    \vspace{-2mm}
\end{figure*}

\subsection{Cross-Modal Re-identification with Multimodal Large Language Models}
The CM-ReID task aims to match identities across diverse modalities like text, sketch, infrared and RGB. IRRA~\cite{jiang2023cross} addresses fine-grained matching in text-based ReID through implicit relation reasoning and a global alignment mechanism. SketchTransformer~\cite{chen2022sketch} proposes an asymmetrical disentanglement learning method based on the Transformer architecture, utilizing dynamic synthesis-assisted sketches to mitigate cross-modal information asymmetry. DEEN~\cite{zhang2023diverse} improves CM-ReID performance under complex lighting conditions through diverse embedding expansion. While effective, this specialized approach has led to a fragmented system. 

To address this, unified frameworks like TriReID~\cite{zhai2022trireid}, AIO~\cite{li2024all}, and FlexiReID~\cite{sun2025flexireid} have demonstrated the feasibility of handling multiple CM-ReID tasks within a single model, motivating the pursuit of more scalable and flexible architectures.

In recent years, MLLMs, which integrate visual encoders with LLMs~\cite{liu2023visual}, have emerged as a powerful tool for ReID due to their strong cross-modal understanding.
One primary application is using MLLMs to understand pedestrian images and producing rich textual descriptions for ReID.
For example, HPMT~\cite{tan2024harnessing} employs CLIP as a backbone and filters potential noise by measuring similarity between generated text and image features. HAM~\cite{jiang2025modeling} introduces style clustering and prompt learning to generate stylistically diverse descriptions, improving generalization in CM-ReID. 
Another line of research reformulates ReID as an interactive task.
ChatReID~\cite{niu2025chatreid} and LLava-ReID~\cite{lu2025llava} explore interactive or multi-round dialogue mechanisms for refining text-based ReID queries. 
However, these methods are often not end-to-end or are confined to a single task, failing to leverage the MLLM as a unified feature extractor.

Recent works such as LVLM-ReID~\cite{wang2024large} and MLLMReID~\cite{yang2024mllmreid} demonstrate the potential of MLLMs or their components as direct feature extractors for single-modal ReID. Despite this promise, their application to a unified, multi-modal CM-ReID setting remains relatively underexplored. 
Furthermore, the massive computational and memory requirements of MLLMs make them infeasible for direct deployment on resource-constrained edge devices. Edge-cloud collaboration~\cite{gu2023ai,wang2024end} is a promising paradigm to address this issue, where computationally intensive inference is performed on cloud servers while lightweight models operate on edge devices for real-time processing. This motivates the need for effective knowledge distillation strategies that can transfer the rich cross-modal understanding capabilities of MLLMs to compact edge-based models suitable for edge deployment in CM-ReID systems.

\subsection{Knowledge Distillation}
Knowledge distillation has been widely adopted in CM-ReID to transfer knowledge from a powerful teacher model to a lightweight student model. 
Early approaches primarily align the output distributions of teacher and student networks using contrastive or classification-based losses~\cite{deng2025dual}. 
Beyond final outputs, several works extend the alignment to intermediate feature representations at key layers~\cite{chen2020maenet,lu2020cross}, or replicate the attention maps of the teacher model to guide the student~\cite{shin2022teaching}. 
Other methods focus on enabling the student to learn relationships between samples of different modalities as captured by the teacher~\cite{chen2022bevdistill}, thereby improving modality-invariant feature learning.

Inspired by~\cite{lee2018self,zhang2024svd}, we explore SVD-based analysis for MLLM distillation in CM-ReID.
Lee et al.~\cite{lee2018self} employ SVD to eliminate spatial redundancy and extract meaningful feature information from feature maps.
SVD-KD~\cite{zhang2024svd} transforms complex tensor-based knowledge into one-dimensional representations via SVD, enabling effective alignment between teacher and student model layers.
Building on these works demonstrating SVD's efficacy in identifying salient features for knowledge distillation, we analyze the cloud model's output through SVD and discover that its feature matrix exhibits low-rank properties.

\section{Method}
\label{sec:method}

\subsection{Model Architectures}
The architecture of our MLLMEmbed-ReID framework is illustrated in ~\Cref{fig:framework}. 
It comprises a cloud-based teacher model for unified cross-modal feature extraction (\emph{i.e.}, RGB, infrared (IR), sketch, and text), and an edge-based student model designed for efficient inference.

%
%
\noindent{\textbf{Cloud-Based Teacher Model}}. 
The cloud-based model is built upon the Qwen2-VL~\cite{wang2025qwen2vl}, serving as a unified feature extractor. To process CM-ReID data, we format the inputs using instructional templates as shown in the upper-left of \Cref{fig:framework}. For instance, a text sample is formatted with the template ``\texttt{The modality of this item: text. The caption is \{\}}'', while visual inputs are accompanied by the task instruction ``\texttt{Generate the image's embedding, focusing on age, gender, clothing, and biometric features.}''
We obtain a sequence of token-wise hidden states $H^c$ from LLM. As the hidden state of the final valid token, $H_{n-1}^c$ encapsulates the semantic information of the entire input sequence, we extract it using the attention mask, where $n$ here is the length of valid token-wise hidden states. It serves as the ReID feature, denoted as $f^c_m \in \mathbb{R}^{d}$, where $m$ represents one of the four modalities (RGB, IR, sketch, text) and $d$ is the feature dimension. This procedure is termed the \textbf{pooling} operation.



\noindent{\textbf{Edge-Based Student Model}}. 
The student model adopts the CLIP (ViT-L/14) architecture~\cite{radford2021cliplearning}. As CLIP is not designed for instruction-based prompting, we input the image and text data without the instructional templates to its respective encoders.
For \textbf{visual modalities} $m \in \{\text{RGB, IR, sketch}\}$, an input image $I_m$ is partitioned into a sequence of $14 \times 14$ patches, prepended with a \texttt{[CLS]} token, and processed by the Vision Transformer. The output hidden state corresponding to the \texttt{[CLS]} token is then passed through a linear projection layer to match the dimensionality of the teacher's feature space. This yields the final visual ReID feature.
For the \textbf{text modality}, an input caption $I_{\text{text}}$ is tokenized and bracketed by \texttt{[BOS]} and \texttt{[EOS]} tokens. The sequence is then processed by the CLIP text encoder, and the resulting representation is similarly passed through a projection layer for dimensional alignment. The final ReID features extracted from the student model are denoted as $f^e_m \in \mathbb{R}^{d}$.

\subsection{Training of Cloud-based Model}
The cloud-based model, possessing pre-trained weights with general knowledge, requires efficient fine-tuning to obtain a model capable of completing unified CM-ReID tasks end-to-end.
Given the strong generalization capabilities of the foundational MLLM, we employ Low-Rank Adaptation (LoRA)~\cite{hu2022lora} for efficient fine-tuning here.
Based on the principle that higher layers capture more task-specific knowledge, we apply LoRA adapters to the final four layers of both the Vision Transformer and the LLM components.
These updated parameters of the LoRA adapter can be represented as $\Delta W\in R^{d \times k}$, where $d$ denotes the input dimension size of the Linear layer and $k$ denotes the output dimension size of the Linear layer. 
Then the LoRA fine-tuning of the cloud-based model can be represented as follows:
\begin{equation}
  W'=W+\Delta W,\space \Delta W=BA,
  \label{eq:lora}
\end{equation}
where $B\in R^{d\times r}$ and $A\in R^{r\times k}$ are trainable low-rank matrices, with rank $r\ll \min(d,k)$.

During the supervised finetuning phase, we train the teacher model with a composite task loss, $\mathcal{L}_{task}$, designed to produce highly discriminative features. This objective combines three distinct loss functions: the ID loss~\cite{mei2024tlreid}, the triplet loss~\cite{zhou2024deep,pham2025scmreid}, and the Similarity Distribution Matching (SDM) loss~\cite{jiang2023cross}. Combining ID Loss and Triplet Loss leverages the strong classification supervision of the former with the discriminative metric learning of the latter, yielding robust and highly separable feature representations for effective ReID.
The ID Loss ($\mathcal{L}_{\text{id}}$) ensures the model accurately distinguishes different identities by minimizing cross entropy between predicted labels and the true identities of pedestrians and can be defined as:
\begin{equation}
\mathcal{L}_{\text{id}} = - \frac{1}{N} \sum_{i=1}^N \log{(\frac{\exp(s_{i,y_i})}{\sum_{j=1}^{C}\exp(s_{i,j})})},
\label{eq:id}
\end{equation}
where $s_{i,j}$ is the score of the $i$-th sample on the $j$-th pedestrian-id (pid) generated by a linear classifier, and $y_i$ is the label of the $i$-th sample.

The Triplet Loss ($\mathcal{L}_{\text{tri}}$) structures the embedding space by minimizing the distance between an anchor feature $f_a$ and a positive feature $f_p$
(same identity), while maximizing the distance to a negative feature $f_n$ (different identity):
\begin{equation}
\mathcal{L}_{\text{tri}} = \frac{1}{N} \sum_{i=1}^N \left[ \|f_a^i - f_p^i\|_2^2 - \|f_a^i - f_n^i\|_2^2 + \alpha \right]_+,
\label{eq:triplet}
\end{equation}
where $\alpha$ is a fixed margin and the function $f(z) = [z]^+ = max(z, 0)$ is the hinge function.

To explicitly enforce robust cross-modal feature alignment and constrain their relative positions within the embedding space, we additionally minimize the SDM loss.
SDM loss ($\mathcal{L}_{sdm}$) aligns the similarity distributions between pairs of modalities. 
Given a mini-batch of $N$ $m$-$n$ pairs, where $m$,$n$ represent different modalities and $N$ represents the batch size, we can construct a set of $m$-$n$ ReID feature pairs $\{(f^c_{i,m},f^c_{j,n}),y_{i,j}\}^N_{j=1}$ for each ReID feature $f_{i,m}^c$ from modality $m$. If $(f^c_{i,m},f^c_{j,n})$ is a matched pair from the same pid, $y_{i,j}=1$. The similarity function is $sim(\textbf{x},\textbf{y})=\textbf{x}^\top\textbf{y}/|\textbf{x}||\textbf{y}|$ and can denote the similarity matrix between $\mathcal{L}_2$ normalized $\textbf{x}$ and $\textbf{y}$. Then the probability of matching pairs can be simply calculated by
\begin{equation}
p_{i,j}=\frac{\exp(sim(f_{i,m}^c,f_{j,n}^c)/\tau)}{\sum_{k=1}^{N}\exp(sim(f_{i,m}^c,f_{k,n}^c)/\tau)},
\label{eq:sdm0}
\end{equation}
where $\tau$ is a temperature hyperparameter which controls the probability distribution peaks. Then the loss from modality $m$ to modality $n$ is computed by
\begin{equation}
\mathcal{L}_{m2n} = \frac{1}{N}\sum_{i=1}^{N}\sum_{j=1}^{N} p_{i,j} \log\left(\frac{p_{i,j}}{q_{i,j} + \epsilon}\right),
\label{eq:sdm1}
\end{equation}
where $q_{i,j}=y_{i,j}/\sum_{k=1}^{N}y_{i,k}$ is the true matching probability.
Meanwhile, conventional CM-ReID tasks typically employ SDM loss to constrain only the modality tested. However, to accurately regularize the vector space, there should be six modality constraints.
\begin{equation}
\mathcal{L}_{SDM} = \sum_{i=1}^6(\mathcal{L}^i_{m2n}+\mathcal{L}^i_{n2m}).
\label{eq:sdm3}
\end{equation}
Finally, the total task loss is the sum of these components:
\begin{equation}
\mathcal{L}_{task} = \mathcal{L}_{id}+\mathcal{L}_{tri}+\mathcal{L}_{SDM}.
\label{eq:total_loss}
\end{equation}

\subsection{Low-Rank Phenomenon}
To develop a more effective distillation strategy than naive feature mimicry, we first analyzed the internal structure of the trained teacher model's feature space. We collected a batch of $n$ ReID feature vectors, forming a feature matrix $F^c_m \in \mathbb{R}^{n \times d}$, where $n$ is the number of samples and $d$ is the feature dimension. We then performed SVD on it:
\begin{equation}
F^c_{m} = U\Sigma V^T,
\label{eq:svd}
\end{equation}
where $U$ and $V$ are $n\times d$ orthogonal matrices. $\Sigma$ is an $n\times d$ diagonal matrix whose diagonal elements are the singular values, $\sigma_k$, ordered by magnitude.

\begin{figure}[t]
    \centering
    \includegraphics[width=\columnwidth]{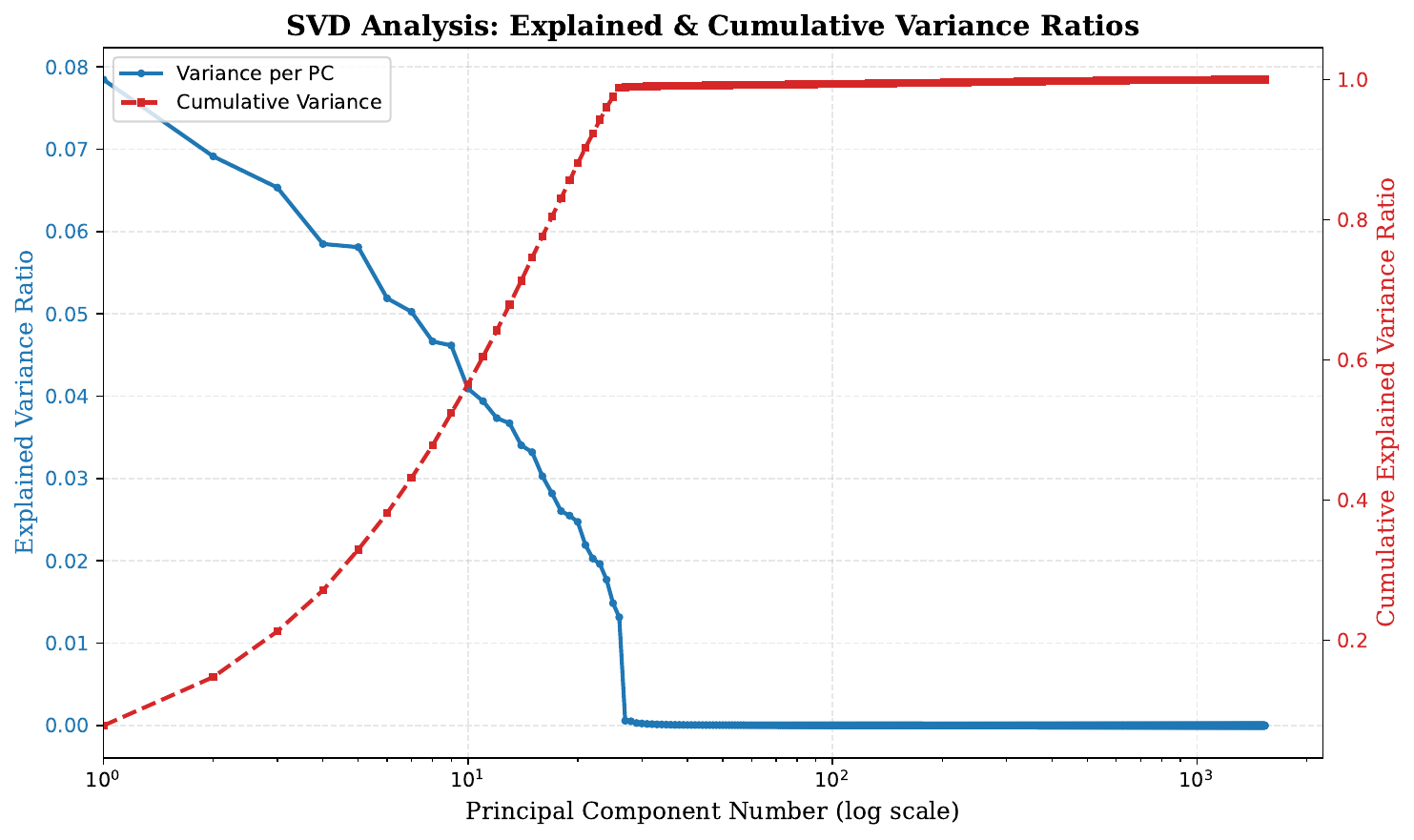}
    \caption{SVD analysis of cloud-based model's ReID feature. The left y-axis shows the explained variance ratio per principal component, while the right y-axis shows the cumulative explained variance ratio. The x-axis is plotted on a logarithmic scale to better visualize the rapid decay of singular values.}
    \label{fig:feat_import}
    \vspace{-3mm}
\end{figure}

As illustrated in Figure~\ref{fig:feat_import}, the analysis revealed a distinct \textbf{low-rank phenomenon}. The singular values exhibit a long-tail distribution, with a small fraction of the principal components accounting for a vast majority of the cumulative variance. This indicates that the most discriminative information is concentrated within a low-dimensional subspace. To quantify the importance of each original feature dimension, we compute a weighted score based on its projection onto the principal components:
\begin{equation}
w_k = \frac{\sigma_k^2}{\sum_{j=1}^{r}\sigma_j^2},
\quad\quad
\text{importance}_i = \sum_{k=1}^{d} |v_{i,k}| w_k,
\label{eq:calimport2}
\end{equation}
where $w_k$ is the normalized variance contribution of the $k$-th principal component, $v_{i,k}$ is the loading of the $i$-th original feature dimension on the $k$-th principal component, $r$ is the rank of $F^c_{m}$ and $d$ is the number of the feature dimension.
These importance scores also follow a long-tail distribution, further motivating a distillation strategy that prioritizes the most significant feature dimensions. 
To further validate the low-rank phenomenon, we conducted additional experiments across diverse datasets, as detailed in \Cref{appendix:robust_lowrank}. The results consistently exhibit the same patterns.

\subsection{Distillation of the Edge-based Model}
To deploy efficient inference models in resource-constrained edge-based computing environments, we designed a basic distillation method and subsequently improved it based on the low-rank phenomenon.
The fundamental distillation method starts with directly employing a cosine comparison loss function to align the ReID feature of the cloud-based model and edge-based model. The cosine loss can be defined as
\begin{equation}
\mathcal{L}_{cosine} = 1-\frac{f^c_mf^e_m}{|f^c_m||f^e_m|}.
\label{eq:cosine_loss}
\end{equation}

According to the low-rank phenomenon, we further designed two loss functions that can rapidly capture highly important features during distillation while also accommodating less significant dimensions. 
First, we design a \textbf{Principal Component Mapping Loss} (PCM) as follows:
\begin{equation}
\mathcal{L}_{pcm} = \mathcal{L}_{match} (f^c_mV^k,f^e_mV^k),
\label{eq:pcmloss}
\end{equation}
where $V^k$ represents the first $k$ principal components extracted from the right singular matrix obtained by performing SVD on the ReID tokens from the cloud-based model and $L_{match}$ is a distillation-based contrastive loss function, such as the cosine loss function. This means that the ReID features of the cloud and edge will be mapped to the vector space containing the top $k$ principal components. In this way, the edge-based model can more readily capture essential dimensional information during the distillation process.

However, ReID falls under embedding-based tasks, and in such tasks, non-essential dimensions also play an equal role in feature matching for retrieval. Therefore, we need to account for non-important dimensions, but directly calculating comparison losses for non-important dimensions would be overly cumbersome. Considering that the low-rank phenomenon we observed reflects the long-tail distribution of important dimensions, the Low-rank approximation theory implies that components corresponding to less important (lower-variance) dimensions can be effectively represented by linear combinations of the principal components. In other words, by learning the relationships between dimensions, we can indirectly account for non-important dimensions. Thus, we designed the \textbf{Feature Relation Loss} (FR) and it can be defined as follows:
\begin{equation}
\mathcal{L}_{fr} = \mathcal{L}_{match} (f^{c\top}_m f^c_m,f^{e\top}_m f^e_m),
\label{eq:frloss}
\end{equation}
where $f^{c\top}_m f^c_m$ is the relationship between features. 

To validate the effectiveness of the distillation method and mitigate catastrophic forgetting during the distillation process, the total distillation loss function is defined as
\begin{equation}
\begin{aligned}
\mathcal{L}_{distill} = & \lambda_{task}{\mathcal L_{task}} + \lambda_{cosine}\mathcal{L}_{cosine} \\
                                                    &+ \lambda_{pcm}\mathcal{L}_{pcm} + \lambda_{fr}\mathcal{L}_{fr},
\end{aligned}
\label{eq:db_loss}
\end{equation}
where $\lambda_{task}$, $\lambda_{cosine}$, $\lambda_{pcm}$ and $\lambda_{fr}$ are the weights of $\mathcal L_{task}$, $\mathcal{L}_{cosine}$, $\mathcal{L}_{pcm}$ and $\mathcal{L}_{fr}$, respectively. 



\section{Experiment}
\label{sec:experiment}

\begin{table*}[t]
\centering
\caption{Comparison with the state-of-the-art methods on QrCM-ReID testing datasets. Rank ($ R $) at $ k $ accuracy (\%) is reported. The best result is displayed in bold, the second-best result is underlined, and the third-best result is italicized.
E2C represents using edge-based features as the query and cloud-based features as the gallery during testing. C2C and E2E follow a similar logic for their respective query and gallery feature types.
Avg. indicates the average value of the \emph{m}AP metric for each method across the three test sets.}
\label{tab:compare}
\vspace{-0.5em}
\resizebox{\textwidth}{!}{
\begin{tabular}{@{}ccccccccccccccccccc@{}}
\toprule
\multirow{2}{*}{Tasks} & \multirow{2}{*}{Methods} & \multirow{2}{*}{Venue} & 
\multicolumn{5}{c}{CUHK-PEDES} & \multicolumn{5}{c}{ICFG-PEDES} & \multicolumn{5}{c}{RSTPReid} & \multirow{2}{*}{Avg.} \\
\cmidrule(lr){4-8} \cmidrule(lr){9-13} \cmidrule(l){14-18}
 & & & R1 & R5 & R10 & \emph{m}AP & \emph{m}INP & R1 & R5 & R10 & \emph{m}AP & \emph{m}INP & R1 & R5 & R10 & \emph{m}AP & \emph{m}INP \\
\midrule
\multirow{7}{*}{$S\to R$ } 
 & Sketch Trans+~\cite{chen2023sketchtrans} & PAMI2023 & 81.39 & 90.61 & 93.54 & 73.72 & 64.72 & 74.83 & 86.75 & 91.52 & 38.64 & 5.68 & 61.37 & 80.15 & 88.29 & 48.94 & 25.73 & 53.77\\
 & DALNet~\cite{liu2024dalnet} & AAAI2024 & 83.03 & 92.39 & 94.58 & 75.39 & 66.82 & 77.28 & 87.84 & 92.61 & 40.35 & 6.18 & 64.68 & 83.27 & 89.06 & 51.08 & 27.13 & 55.61\\
 & UniReID~\cite{chen2023unireid} & CVPR2023 & 84.87 & - & - & 78.85 & 68.55 & 77.47 & - & - & 40.41 & 6.31 & 65.80 & - & - & 51.22 & 27.47 & 56.83\\
 & FlexiReID~\cite{sun2025flexireid} & ICML2025 & 84.92 & 93.17 & 95.02 & 79.21 & 68.83 & 79.28 & 89.69 & 93.37 & 41.21 & 6.85 & 66.79 & 84.52 & 90.39 & 52.72 & 28.36 & 57.71\\
 & MLLMEmbed-ReID(Ours, C2C) & - & \textbf{92.87} & \textbf{97.27} & \textbf{98.42} & \textbf{89.82} & \textbf{84.21} & \underline{87.36} & \underline{94.05} & \underline{95.63} & \textbf{60.39} & \textbf{20.47} & \underline{86.15} & \textit{94.56} & \textit{96.19} & \underline{74.67} & \underline{54.51} & \textbf{74.96}\\
 & MLLMEmbed-ReID(Ours, E2E) & - & \underline{89.14} & \underline{95.97} & \underline{97.55} & \underline{85.23} & \textit{77.64} & \textbf{88.10} & \textbf{94.69} & \textbf{96.25} & \underline{58.90} & \underline{19.44} & \textbf{87.01} & \underline{95.64} & \textbf{96.91} & \textbf{75.82} & \textbf{56.53} & \underline{73.32}\\
 & MLLMEmbed-ReID(Ours, E2C) & - & \textit{87.13} & \textit{95.76} & \textit{97.48} & \textit{84.84} & \underline{78.27} & \textit{81.37} & \textit{92.28} & \textit{94.75} & \textit{54.90} & \textit{17.70} & \textit{84.04} & \textbf{95.74} & \underline{96.75} & \textit{73.91} & \textit{53.76} & \textit{71.22}\\
\hline
\multirow{6}{*}{$IR\to R$ } 
 & GUR~\cite{yang2023gur} & ICCV2023 & 82.06 & 91.72 & 93.95 & 75.84 & 66.86 & 80.31 & 90.89 & 92.78 & 44.36 & 6.90 & 73.42 & 86.29 & 91.35 & 60.43 & 38.52 & 60.21\\
 & SDCL~\cite{yang2024sdcl} & CVPR2024 & 84.57 & 92.73 & 94.58 & 77.32 & 68.20 & 81.36 & 91.83 & 94.07 & 45.81 & 7.92 & 74.67 & 87.94 & 93.16 & 62.75 & 39.93 & 61.96\\
 & FlexiReID & ICML2025 & 85.26 & 93.25 & 95.31 & 79.43 & 69.39 & 82.03 & 92.19 & 94.27 & 46.76 & 8.47 & 75.36 & 88.71 & 93.27 & 63.22 & 40.82 & 63.14\\
 & MLLMEmbed-ReID(Ours, C2C) & - & \textbf{92.87} & \textbf{97.84} & \textbf{98.62} & \textbf{90.28} & \textbf{85.07} & \underline{86.77} & \underline{94.23} & \underline{95.84} & \underline{59.34} & \underline{19.43} & \underline{85.81} & \textbf{95.28} & \underline{97.06} & \underline{72.91} & \textit{52.06} & \textbf{74.18}\\
 & MLLMEmbed-ReID(Ours, E2E) & - & \underline{90.45} & \underline{97.04} & \textit{97.95} & \underline{87.19} & \textit{80.27} & \textbf{88.95} & \textbf{95.36} & \textbf{96.70} & \textbf{59.94} & \textbf{19.94} & \textbf{85.89} & \textit{94.95} & \textbf{97.10} & \textbf{75.21} & \textbf{55.63} & \underline{74.11} \\
 & MLLMEmbed-ReID(Ours, E2C) & - & \textit{89.00} & \textit{96.61} & \underline{98.18} & \textit{86.74} & \underline{80.62} & \textit{82.81} & \textit{93.25} & \textit{95.59} & \textit{56.08} & \textit{18.30} & \textit{85.01} & \underline{95.17} & \textit{96.97} & \textit{73.31} & \underline{52.25} & \textit{72.04} \\
\hline
\multirow{14}{*}{$T\to R$ } 
 & ViTAA~\cite{wang2020vitaa} & ECCV2020 & 55.97 & 75.84 & 83.52 & - & - & 50.98 & 68.79 & 75.78 & - & - & - & - & - & - & - & -\\
 & DSSL~\cite{zhu2021dssl} & MM2021 & 59.98 & 80.41 & 87.56 & - & - & - & - & - & - & - & 32.43 & 55.08 & 63.19 & - & - & -\\
 & LBUL+BERT~\cite{wang2022lbul} & MM2022 & 64.04 & 82.66 & 87.22 & - & - & - & - & - & - & - & 45.55 & 68.20 & 77.85 & - & - & -\\
 & SAF~\cite{li2022saf} & ICASSP2022 & 64.13 & 82.62 & 88.40 & 58.61 & - & 54.86 & 72.13 & 79.13 & 32.76 & - & 44.05 & 67.30 & 76.25 & 36.81 & - & 42.73\\
 & TIPCB~\cite{chen2024tipcb} & NeurIPS2022 & 64.26 & 83.19 & 89.10 & - & - & 54.96 & 74.72 & 81.89 & - & - & - & - & - & - & - & -\\
 & CAIBC~\cite{wang2022caibc} & MM2022 & 64.43 & 82.87 & 87.35 & - & - & - & - & - & - & - & 47.35 & 69.55 & 79.00 & - & - & -\\
 & LGUR~\cite{shao2022lgur} & MM2022 & 65.25 & 83.12 & 89.00 & - & - & \textit{59.02} & 75.32 & 81.56 & - & - & - & - & - & - & - & -\\
 & LVT~\cite{shu2022lvt} & ECCV2022 & 65.59 & 83.11 & 89.21 & - & - & 56.04 & 73.60 & 80.22 & - & - & 46.70 & 70.00 & 78.80 & - & - & -\\
 & UNIReID & CVPR2023 & \textit{68.71} & \textit{85.93} & \textit{90.84} & - & - & 61.28 & \underline{77.40} & \textit{83.16} & - & - & \textbf{60.25} & 79.85 & \textit{87.10} & - & - & -\\
 & CSKT~\cite{liu2024cskt} & CVPR2023 & \textbf{69.70} & \textbf{86.92} & \textbf{91.80} & \underline{62.74} & - & 58.90 & \textit{77.31} & \underline{83.56} & 33.87 & - & \textit{57.75} & \textbf{81.30} & \textbf{88.35} & 46.43 & - & \textit{47.68} \\
 & FlexiReID & ICML2025 & \underline{69.20} & \underline{86.43} & \underline{91.41} & \textit{62.47} & \underline{48.32} & \textbf{61.34} & \textbf{78.41} & \textbf{83.92} & 35.73 & 7.53 & 55.79 & \textit{79.62} & 86.48 & 45.37 & 26.25 & \underline{47.86} \\
 & MLLMEmbed-ReID(Ours, C2C) & - & 65.29 & 82.22 & 87.56 & \textbf{62.92} & \textbf{52.11} & \underline{59.49} & 76.36 & 82.07 & \textbf{41.52} & \textbf{12.10} & \underline{59.28} & \underline{79.89} & \underline{87.30} & \textbf{50.96} & \textbf{31.76} & \textbf{51.80}\\
 & MLLMEmbed-ReID(Ours, E2E) & - & 57.29 & 75.47 & 82.90 & 55.65 & 44.49 & 54.35 & 70.64 & 77.03 & \textit{37.46} & \textit{11.15} & 54.01 & 74.12 & 83.90 & \underline{47.29} & \underline{30.72} & 46.80 \\
 & MLLMEmbed-ReID(Ours, E2C) & - & 58.80 & 77.35 & 84.98 & 58.03 & \textit{47.87} & 52.79 & 71.32 & 77.98 & \underline{38.18} & \underline{11.68} & 52.79 & 74.18 & 82.89 & \textit{46.44} & \textit{28.82} & 47.55\\
\bottomrule
\end{tabular}
}
\end{table*}

\subsection{Experimental Setup}
\textbf{CM-ReID Dataset.} In this work, we will follow the approach outlined in \cite{sun2025flexireid,zhai2022trireid} and introduce three text-based ReID datasets CUHK-PEDES\cite{ding2021semantically}, ICFG-PEDES\cite{shen2023pedestrian}, and RSTPReid\cite{zhu2021dssl}. Subsequently, since these datasets already possess RGB and text modalities, we apply \cite{zhu2023stylegan3} to generate sketch modalities as described in \cite{chen2022sketch}. Subsequently, we will apply \cite{ozkanouglu2022infragan} to generate infrared-visible-light image modalities. This model is trained using visible and infrared-visible image pairs, which enables it to effectively capture thermal radiation information. This modality expansion method and the processed dataset have been proven effective in relevant papers, and we refer to this expanded dataset after modal expansion as the Quadruple Cross Modal Re-identification (QrCM-ReID) dataset. By augmenting additional modalities, we construct multi-modal pedestrian data to simulate real-world conditions, enabling the ReID model to jointly address multiple CM-ReID tasks.

\textbf{Evaluation Protocols.} We follow standard CM-ReID evaluation protocols, using Rank-n accuracy, mean Average Precision (\emph{m}AP), and mean Inverse Negative Penalty (\emph{m}INP) to evaluate every model in MLLMEmbed-ReID.

\textbf{Implementation Details.} For the cloud model, we apply Qwen2-VL-2B as our backbone network. Throughout training, LoRA rank is set to 16, and the $\alpha$ is set to 2.
We perform experiments on four NVIDIA 3090 GPUs. For edge-based models, we adopt CLIP(ViT-L/14) as our backbone network. A linear mapping layer is configured for each modality to transform vectors from the edge-based model space into a vector space aligned with the cloud-based model. The three visual modalities (RGB, IR, sketch) share a Vision Transformer visual encoder, while the text modality uses a Transformer text encoder. All parameters of the cloud-based model are frozen throughout the distillation and training process while those of the edge-based model are unfrozen. We choose the cosine loss function as $\mathcal{L}_{match}$.
Each identity includes at least two modality groups per batch, totaling eight samples. The batch size is set to 32 per GPU, and we use four GPUs, resulting in a total batch size of 128. The images are resized to 280$\times$140 to fit the MLLM backbone of the cloud-based model. The fine-tuned cloud-based model uses a fixed text length of 160, while the distilled edge-based model employs a fixed text length of 77 because the former has longer prompts. The AdamW optimizer is uniformly adopted. Fine-tuning experiments run for 120 epochs, while distillation runs for 60 epochs. The initial learning rate is set to 1e-5 and is decayed to a minimum learning rate of 1e-6 using a cosine scheduler. The $k$ of PCM is a hyperparameter, with a default value of 50.




\subsection{Performance Comparison}

\textbf{Cloud-based Model Performance}. In our experiment, we first finetune the cloud-based model using the QrCM-ReID training dataset, then distill it to obtain an edge-based model. Finally, we evaluated the model's performance when queries and galleries originated from different models to simulate real-world scenarios. Our models are tested on three cross-modal tasks: $IR\to R$, $T\to R$, and $S\to R$. \Cref{tab:compare} presents the performance of cloud-based and edge-based models on the QrCM-ReID test dataset. Currently, there is little work on unified CM-ReID. Only FlexiReID~\cite{sun2025flexireid} and our work have simultaneously completed tests for three tasks. Other methods cannot jointly process the four modalities in the QrCM-ReID dataset.

For tasks $S\to R$ and $IR\to R$, our cloud-based model achieved state-of-the-art performance on all metrics across the three test sets. For example, in task $IR\to R$, our cloud model achieved R1 of 92.87 and \emph{m}AP of 90.28 on the CHUK-PEDES dataset. In the ICFG-PEDES dataset, it achieved R1 of 86.77 and \emph{m}AP of 59.34. In the RSTPReid dataset, R1 reached 85.81 and \emph{m}AP reached 72.91.

\begin{figure}[t]
    \centering
    \includegraphics[width=\columnwidth]{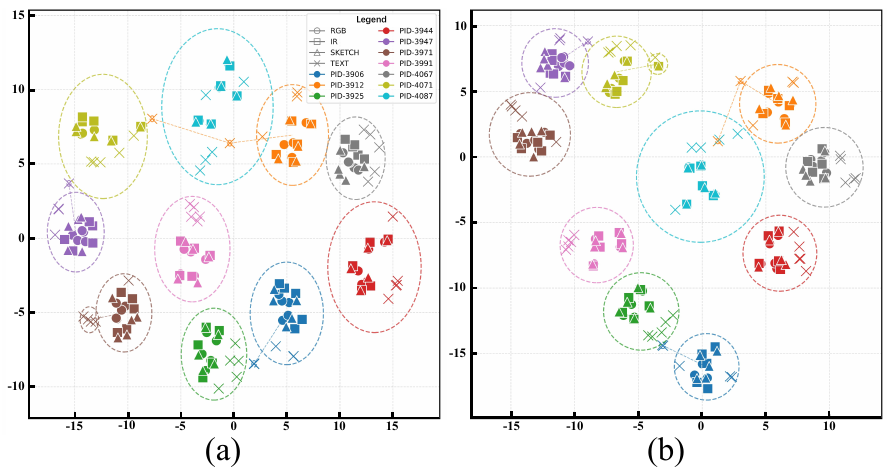}
    \caption{(a) and (b) represent the t-SNE visualization of the cloud-based model and edge-based model, respectively. Scatter points of different shapes represent different modal data. Different scatter colors represent different pedestrian IDs.}
    \label{fig:tsne}
    \vspace{-3mm}
\end{figure}
\begin{figure}[t]
    \centering
    \includegraphics[width=\columnwidth]{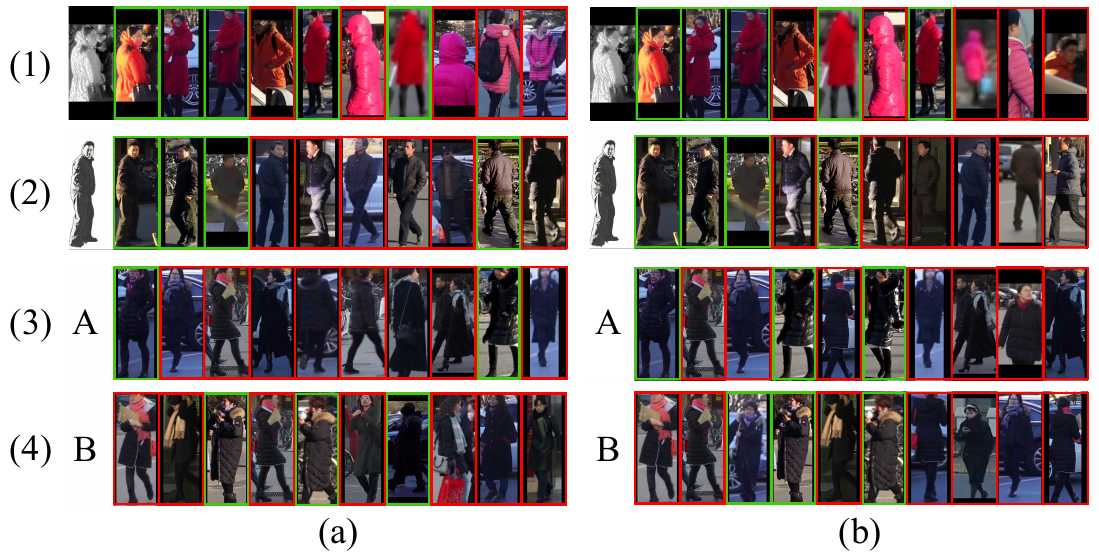}
    \caption{(a) and (b) represent the recognition results of the cloud-based model and edge-based model, respectively. Images are from the QrCM-ReID dataset. (1) and (2) represent the IR$\to$R and S$\to$R tasks, respectively, while (3) and (4) correspond to the T$\to$R task. Caption A: This woman is a ponytail, wearing a black down jacket, black trousers, black boots, wearing a purple scarf and glasses. She walks with her hand in her pocket. Caption B: This woman is wearing a black coat, black trousers and black shoes. She was wearing glasses and a purple scarf. She walks while watching her cell phone.}
    \vspace{-1.5em}
    \label{fig:case}    
\end{figure}

In $T \to R$, our cloud-based model achieved SOTA state-of-the-art levels on \emph{m}AP, while Rank-n accuracy approaches state-of-the-art levels.
As shown in the \Cref{fig:tsne} and \Cref{fig:case}, through t-SNE and case analysis, we find that the cloud-based model primarily lacks the ability to perceive local information due to the absence of customized modules like Mixture of Experts (MoE) in FlexiReID~\cite{sun2025flexireid}.
However, higher \emph{m}AP performance demonstrates that our cloud-based model exhibits stronger comprehensive retrieval capabilities and higher generalization performance but lacks fine-grained constraints on the most accurate retrieval target. 

\textbf{Edge-based Model Performance}. Our edge-based model delivers CM-ReID performance on par with cloud-based MLLM. For tasks $S\to R$ and $IR\to R$, the edge-based model continues to achieve state-of-the-art performance. For instance, in task $S\to R$, the edge-based model achieves R1 of 89.14 and \emph{m}AP of 85.23 on CUHK-PEDES. On the ICFG-PEDES dataset, it achieved R1 of 88.10 and \emph{m}AP of 58.90. On the RSTPReid dataset, R1 achieved 87.01 and \emph{m}AP reached 75.82. 
For task $T\to R$, the edge-based model surpassed all existing methods in \emph{m}AP on both ICFG-PEDES and RSTPReid, while its rank-n accuracy closely approached the state-of-the-art.

\begin{table*}[t]
\centering
\caption{Ablation study about each component of edge-based Model distillation on CUHK-PEDES. Avg. here represents the average \emph{m}AP across all experiments for the three tasks in CUHK-PEDES.}
\label{tab:ablation}
\resizebox{\textwidth}{!}{
\begin{tabular}{@{}lccccccccccccccccccc@{}}
\toprule
\multirow{2}{*}{Components} & \multirow{2}{*}{cosine} & \multirow{2}{*}{PCM} & \multirow{2}{*}{FR} &
\multicolumn{5}{c}{$IR\to R$} & \multicolumn{5}{c}{$T\to R$} & \multicolumn{5}{c}{$S\to R$} & \multirow{2}{*}{Avg.} \\
\cmidrule(lr){5-9} \cmidrule(lr){10-14} \cmidrule(l){15-19}
& & & & R1 & R5 & R10 & \emph{m}AP & \emph{m}INP & R1 & R5 & R10 & \emph{m}AP & \emph{m}INP & R1 & R5 & R10 & \emph{m}AP & \emph{m}INP \\
\midrule
 (a) Edge-training-only & & & & 83.63 & 93.92 & 96.69 & 80.90 & 73.00 & 46.08 & 68.11 & 77.89 & 46.98 & 36.42 & 84.33 & 93.74 & 95.89 & 80.67 & 72.44 & 69.50\\
 (b) (a)+cosine & \checkmark($\lambda=0.99$) & & & 89.34 & 96.17 & 97.51 & 85.58 & 78.06 & 49.50 & 67.94 & 76.38 & 48.43 & 37.99 & 87.16 & 95.26 & 97.14 & 83.28 & 75.35 & 72.43\\
 (c) (b)+FR & \checkmark($\lambda=0.29$) & \checkmark($\lambda=0.70$) & & 90.52 & 96.96 & 98.14 & 87.42 & 80.68 & 57.22 & \textbf{75.46} & \textbf{83.12} & 55.63 & 44.33 & \textbf{89.39} & 95.97 & 97.4 & 85.42 & 77.91 & 76.16\\
 (d) (b)+PCM & \checkmark($\lambda=0.29$) & \checkmark($\lambda=0.70$) & & \textbf{90.97} & \textbf{97.30} & \textbf{98.26} & \textbf{87.71} & \textbf{80.97} & 57.01 & 75.12 & 82.34 & \textbf{55.70} & \textbf{44.89} & 89.31 & \textbf{96.37} & \textbf{97.66} & \textbf{85.56} & \textbf{78.19} & \textbf{76.26}\\
 (e) (b)+PCM+FR & \checkmark($\lambda=0.29$) & \checkmark($\lambda=0.35$) & \checkmark($\lambda=0.35$) & 90.45 & 97.04 & 97.95 & 87.19 & 80.27 & \textbf{57.29} & 75.47 & 82.90 & 55.65 & 44.49 & 89.14 & 95.97 & 97.55 & 85.23 & 77.64 & 76.02\\
\bottomrule
\end{tabular}
}
\vspace{-2mm}
\end{table*}

\begin{table}[t]
\centering
\caption{Ablation Study on LoRA rank and target module parameters $\alpha$ on CUHK-PEDES. Avg. here represents the average \emph{m}AP across all experiments for the three tasks in CUHK-PEDES.}
\label{tab:lora_config}
\resizebox{0.4\columnwidth}{!}{
\begin{tabular}{@{}cccc@{}}
\toprule
LoRA Rank & $\alpha$ & Avg. \\
\midrule 
8 & 1 & 69.41\\
8 & 2 & 66.92\\
8 & 4 & 60.01\\
16 & 1 &77.26\\
16 & 2 & \textbf{80.87} \\
16 & 4 & 78.69\\
\bottomrule
\end{tabular}
}
\end{table}

\subsection{Ablation Study}

To observe the effects of different distillation functions during the edge-based model distillation process, the results of ablation experiment are shown as the \Cref{tab:ablation}. To prevent the task loss from dominating and potentially perturbing the distillation dynamics, we maintain $\lambda_{task}=0.01$ at a significantly low scale. We normalize the distillation loss weights such that $\lambda_{task}+\lambda_{cosine}+\lambda_{pcm}+\lambda_{fr}=1$, which ensures a stable gradient scale during the convergence process. 
As illustrated by the partial tuning results in \Cref{appendix:distill_obj_coef}, the current design is the culmination of extensive hyperparameter optimization.

In configuration (a), the edge-based model is trained without the distillation method.
Configuration (b) employs a traditional alignment method, aligning the ReID feature of the edge-based model with that of the cloud-based model using a cosine similarity loss function. 
Configuration (c) and (d) incorporate FR and PCM respectively, building upon the foundation of configuration (b). 
Configuration (e) simultaneously employs both PCM and FR to investigate their performance when assigned equal weights.
We assign an extremely low weight of $\lambda$=0.01 to $\mathcal L_{task}$. The sum of the other distillation loss weights should be 0.99, where the baseline cosine loss weight is set to 0.29, and both weights of FR and PCM are 0.35. This approach prevents catastrophic forgetting during the distillation.
As we anticipated, configuration (c) and (d) confirmed that both losses derived from our discovery of low-rank linearity indeed deliver more efficient distillation results. Configuration (e) shows PCM and FR significantly improve the performance of the $T\to R$ task.

However, it is worth noting that when PCM and FR are combined, as demonstrated in configuration (e), no significant improvement was achieved compared to configuration (c) and (d), and even a slight decline was observed. 
As shown in \Cref{appendix:mutual_fr_pcm}, although combining the two distillation loss functions leads to a slight decrease in the testing metrics, it significantly enhances training efficiency.
Furthermore, the observed performance decline suggests a conflict between PCM's targeted alignment and FR's holistic regularization. While PCM focuses the student on the low-rank subspace containing the principal components, FR enforces mimicry of the entire feature correlation matrix. This creates competing signals for the lightweight student, potentially forcing a sub-optimal trade-off that slightly compromises the learning of the most salient features.
\vspace{-2mm}
\subsection{Hyper-Parameter Analysis}

To precisely control the fine-tuning depth and evaluate its impact, we varied the LoRA rank (using values of 8 and 16) and the selection of target modules for adaptation.

As detailed in \Cref{sec:method}, we implemented a hierarchical LoRA strategy. Adapters were applied densely to the final 4 layers of the Vision Transformer and LLM, and more sparsely to every 
$\alpha$-th preceding layer ($\alpha\in\{1,2,4\}$). This approach focuses adaptation on task-specific upper layers while preserving the model's foundational pre-trained knowledge in the lower layers.

The results in \Cref{tab:lora_config} validate this strategy, with the optimal performance on CUHK-PEDES achieved at a LoRA Rank of 16 and $\alpha=2$. Notably, the performance degrades when $\alpha=1$, indicating that an overly dense application of LoRA adapters is counterproductive, likely by disrupting valuable pre-trained features in the lower layers.
\vspace{-2mm}

\section{Conclusion}
\label{sec:conclusion}
We present MLLMEmbed-ReID, a unified framework for cross-modal ReID that achieves state-of-the-art performance by adapting a MLLM after a hierarchical LoRA SFT as a powerful teacher.
To bridge this model to resource-constrained devices, we introduce an efficient distillation strategy based on the key insight that the teacher's feature space exhibits a low-rank property.
Validated on multiple challenging benchmarks, our method delivers a highly practical and scalable cloud-edge pipeline for deploying MLLM-level intelligence efficiently on edge devices, setting a new paradigm for cloud-edge, versatile, and MLLM-based CM-ReID.

\section*{Impact Statement}

This paper presents work aimed at advancing the field of Machine Learning, particularly in cross-modal representation learning for person re-identification (ReID). 
Our framework has potential applications in intelligent surveillance and public safety. 
We acknowledge that such technologies carry inherent ethical considerations, including privacy concerns and the potential for algorithmic bias. 
These implications are well-established in the deployment of AI-powered identification systems. 
While these aspects are critical, they align with the broader discourse in the field; thus, we do not believe a substantial discussion is required here beyond this acknowledgment. 
We advocate for the responsible development and ethical deployment of ReID technologies.


\bibliography{example_paper}
\bibliographystyle{icml2026}

\newpage
\appendix
\onecolumn

\section{Impact of Distillation Objective Coefficients}
\label{appendix:distill_obj_coef}
We have already mentioned that we maintain $\lambda_{task}=0.01$ at a significantly low scale and $\lambda_{task}+\lambda_{cosine}+\lambda_{pcm}+\lambda_{fr}=1$.
Then, we design 3 configurations for the ratio of $\lambda_{cosine}$ and $(\lambda_{pcm} + \lambda_{fr})$, denoted as (a)0.70:0.29;(b)0.495:0.495;(c)0.29:0.70. Across all configurations, we maintain equal weights for $\lambda_{pcm}$ and $\lambda_{fr}$, reflecting our premise that these two distillation objectives are of equivalent importance in reshaping the model's feature space. 

As reported in \cref{tab:abl_loss_analysis}, the experimental results on CUHK-PEDES demonstrate that (c) achieves the superior performance across all evaluation metrics. Compared with (a), which prioritizes global semantic alignment, (c) effectively captures the fine-grained low-rank features from the MLLM by significantly increasing the weighting intensity of the structural distillation components. This leads to a substantial improvement in Rank-1 accuracy and mAP, thereby validating the superiority of the current 29:70 ratio design.

\begin{table*}[t]
\centering
\caption{Sensitivity analysis of distillation objective coefficients on the CUHK-PEDES dataset. (c) allocates more weight to  distillation ($L_{pcm}$ and $L_{fr}$), achieving the best performance.}
\label{tab:abl_loss_analysis}
\resizebox{\textwidth}{!}{
\begin{tabular}{@{}ccccccccccccccccc@{}}
\toprule
\multirow{2}{*}{Components} &
\multicolumn{5}{c}{$IR\to R$} & \multicolumn{5}{c}{$T\to R$} & \multicolumn{5}{c}{$S\to R$} & \multirow{2}{*}{Avg.} \\
\cmidrule(lr){2-6} \cmidrule(lr){7-11} \cmidrule(l){12-16}
& R1 & R5 & R10 & \emph{m}AP & \emph{m}INP & R1 & R5 & R10 & \emph{m}AP & \emph{m}INP & R1 & R5 & R10 & \emph{m}AP & \emph{m}INP \\
\midrule
 (a)       & 88.37 & 95.80 & 97.48 & 84.54 & 76.79 & 51.92 & 66.10 & 73.56 & 46.94 & 36.68 & 86.49 & 95.06 & 96.91 & 82.54 & 74.43 & 71.34\\
 (b)       & 89.71 & 96.91 & 97.98 & 86.65 & 79.79 & 53.16 & 72.68 & 80.54 & 52.45 & 41.87 & 88.27 & 95.87 & 97.31 & 84.49 & 76.78 & 74.53\\
 (c)       & 90.45 & 97.04 & 97.95 & 87.19 & 80.27 & 57.29 & 75.47 & 82.90 & 55.65 & 44.49 & 89.14 & 95.97 & 97.55 & 85.23 & 77.64 & \textbf{76.02}\\
\bottomrule
\end{tabular}
}
\end{table*}

\section{Convergence Analysis: Mutual Reinforcement of Distillation Objectives}
\label{appendix:mutual_fr_pcm}
\begin{figure*}
  \centering
   \includegraphics[width=\linewidth]{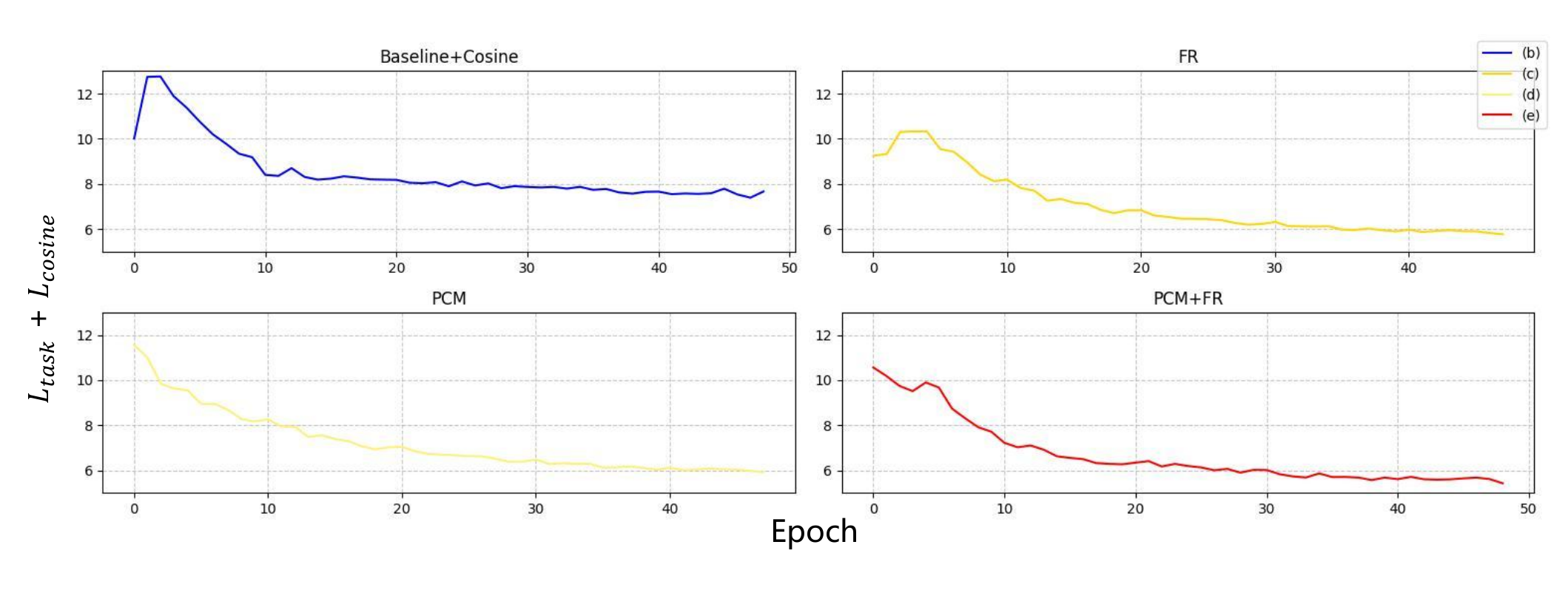}
   \caption{Convergence analysis of different distillation configurations. The Y-axis represents the shared loss ($L_{task} + L_{cos}$) to ensure a fair comparison across all schemes. Our joint configuration (e) demonstrates better optimization stability and a faster convergence rate.}
   \label{fig:training_curve}
   \vspace{-2mm}
\end{figure*}

As shown in row (e) of \cref{tab:ablation}, no significant improvement was achieved compared to configuration (c) and (d), and even a slight decline was observed. Although the joint effect of the two distillation loss functions does not yield a substantial further improvement in final accuracy, as illustrated in \cref{fig:training_curve}, the combination of $L_{pcm}$ and $L_{uni}$ (FR) significantly enhances the training dynamics. Specifically, "PCM+FR" exhibits a more stable convergence trajectory with fewer initial fluctuations and achieves a faster reduction in the shared objective compared to the individual components. We select the sum of task loss and cosine loss as the evaluation metric for this analysis because they represent the primary optimization targets common to all configurations, providing a fair and consistent baseline to observe how auxiliary structural distillation objectives facilitate the overall learning process.

\section{Generality of the Low-Rank Phenomenon}
\label{appendix:robust_lowrank}
\begin{figure*}
  \centering
   \includegraphics[width=\linewidth]{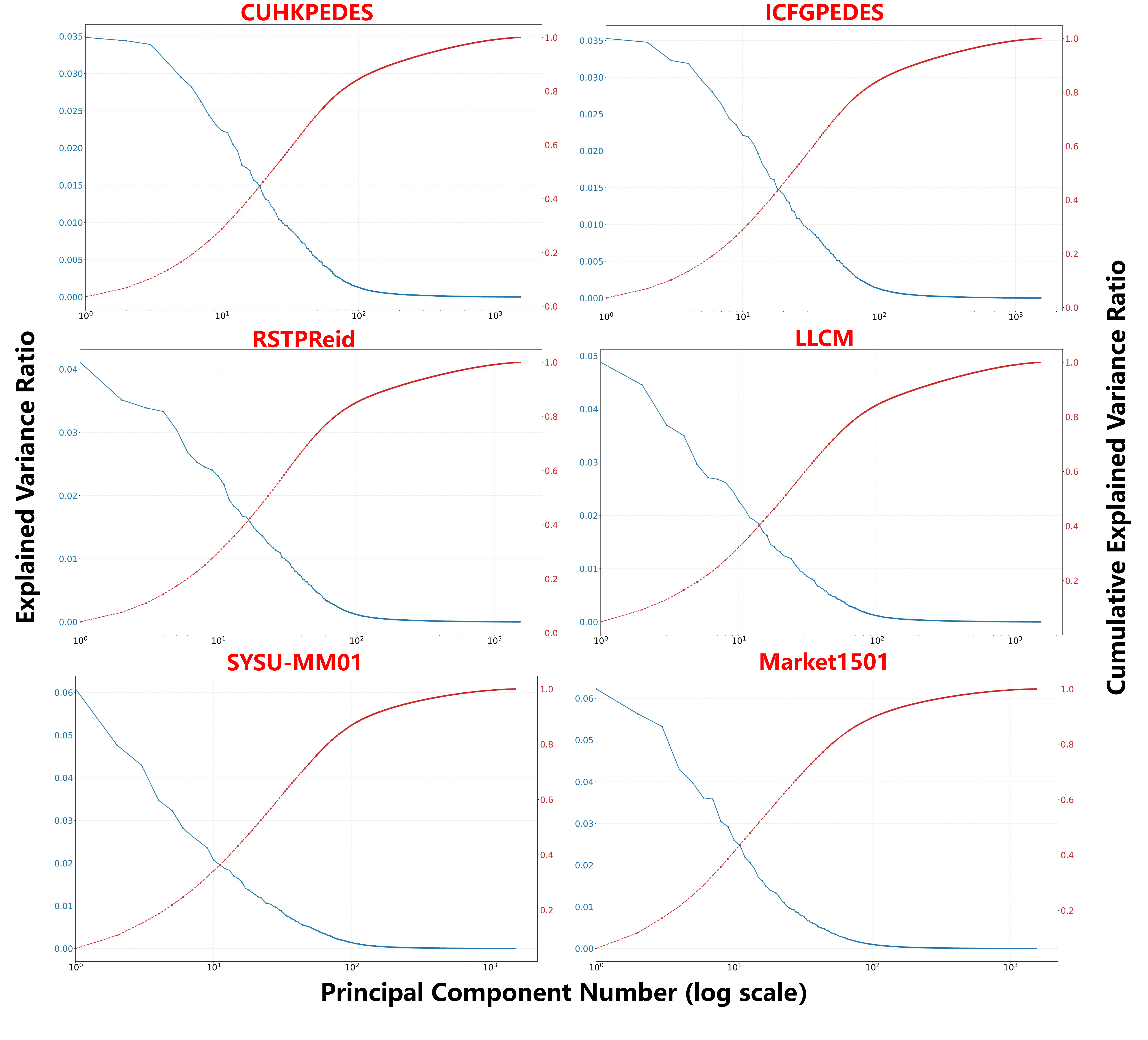}
   \caption{We present the explained variance ratio (blue curves) and cumulative explained variance ratio (red curves) on diverse data distributions. The consistent decay patterns across all trials demonstrate that the observed low-rank structure is a fundamental and dataset-agnostic property of MLLM embeddings.}
   \label{fig:lowrank_diffds}
   \vspace{-2mm}
\end{figure*}

To further validate the low-rank phenomenon, we conduct extensive experiments across multiple independent datasets including CUHKPEDES, ICFGPEDES, RSTPReid, LLCM~\cite{zhang2023diverse}, SYSU-MM01~\cite{zheng2025sysu} and Market1501~\cite{fouad2025market}.
Following the exact experimental protocol described in Section \Cref{sec:method}, we perform 4 randomized trials for each dataset to eliminate the potential bias of specific samples.
As illustrated in \Cref{fig:lowrank_diffds}, the singular value distributions across diverse datasets consistently mirror the patterns observed in \Cref{fig:feat_import}. 
Due to the high consistency observed across the four randomized trials, we present only one representative result in the visualization for conciseness and better layout.
Despite the variations in data domain and distribution, the inherent low-rank structure remains remarkably stable. 
This empirical evidence suggests that the low-rank phenomenon is a fundamental characteristic of MLLM embeddings, rather than an artifact of a specific dataset or sampling noise. 

\section{Edge Device Deployment Performance Validation}

To address potential concerns regarding the practical feasibility of edge deployment, we conducted a performance evaluation on TWOWIN TW-T208. The experimental setup includes: GPU acceleration, bfloat16 precision, image size of 392×140 pixels, and Flash Attention2 disabled. We evaluated our distilled edge model under two scenarios: real-time single-sample processing (BatchSize=1) and batch processing (BatchSize=4). The results of the report are shown in \Cref{tab:edge_deployment}

\begin{table}[h]
\centering
\caption{Inference Performance (Latency and Throughput) report of the edge model on TWOWIN TW-T208. The evaluation metrics, latency and throughput, are measured in milliseconds (ms) and Samples Per Second (SPS), respectively.}
\label{tab:edge_deployment}
\small
\begin{tabular}{lccccc}
\toprule
{Running Type} & {Batch Size} &  {Latency (ms)} & {Throughput (SPS)} \\
\midrule
Real-time Processing & 1 & 194.73 & 5.14 \\
Batch Processing & 4 & 576.67 & 6.94 \\
\bottomrule
\end{tabular}
\end{table}


\end{document}